\documentclass[lettersize,journal]{IEEEtran}
\usepackage{amsmath,amsfonts}
\usepackage{algorithmic}
\usepackage{algorithm}
\usepackage{array}
\usepackage[caption=false,font=normalsize,labelfont=sf,textfont=sf]{subfig}
\usepackage{textcomp}
\usepackage{stfloats}
\usepackage{url}
\usepackage{verbatim}
\usepackage{graphicx}
\usepackage{bm}

\usepackage{caption} %

\usepackage{cite}
\usepackage{booktabs}   %
\usepackage{multirow}   %
\usepackage{caption}    %

\DeclareCaptionLabelSeparator{period}{. }
\usepackage{makecell}

\usepackage{amssymb}
\usepackage{pifont}
\newcommand{\cmark}{\checkmark} 

\newcommand{\xmark}{\ding{53}}

\begin{document}

\title{VISTA: \textbf{V}isually \textbf{I}nferred \textbf{S}patial Con\textbf{T}act
\textbf{A}ttention\\
for Contact-Rich Manipulation}

\author{
Jiayi Chen$^{1,*}$,
Wenlong Dong$^{2,*}$,
Yan Huang$^{1,*}$,
Xianglin Chen$^{1}$,
Zijian Lin$^{1}$,
Jiaqi Yin$^{3}$,
Yushan Liu$^{1}$,\\
Wenbo Ding$^{1,\dagger}$
\thanks{
$^{*}$These authors contributed equally to this work.
$^{1}$Tsinghua Shenzhen International Graduate School, Tsinghua University, Shenzhen, China.
$^{2}$Department of Electronic and Electrical Engineering, Southern University of Science and Technology, Shenzhen, China.
$^{3}$School of Future Technology, Harbin Institute of Technology, Harbin, China.
$^{\dagger}$Corresponding author: Wenbo Ding (e-mail: ding.wenbo@sz.tsinghua.edu.cn).
}
}

\maketitle

\begin{abstract}

Contact-rich manipulation requires precise interaction feedback. While vision-centric imitation learning is prevalent, external visual observations provide indirect and ambiguous cues about contact states, particularly under occlusion or subtle object--gripper interactions; dedicated tactile or force sensors can provide rich contact information but introduce additional hardware complexity, calibration requirements, and deployment costs. To bridge this gap, we propose VISTA-Policy, an imitation learning paradigm that utilizes the Visual Deformation Field (VDF), a 3D displacement representation of a compliant gripper, as high-dimensional visuo-physical feedback. The framework integrates: 1) a Physics-Aware Encoding Engine for real-time VDF decoding; 2) an Energy Aggregation Denoising Mechanism to isolate true interaction signals; and 3) a Deformation-Augmented Policy Network with incremental gripper actions for precise closed-loop correction. Extensive evaluations on Cross-Scale Object Grasping, Cap Unscrewing, and Calligraphy Writing demonstrate that VISTA-Policy outperforms the strong pure-vision baseline 3D Diffusion Policy and the tactile baseline. VISTA-Policy further demonstrates substantial out-of-distribution generalization to unseen object scales and robustness against dynamic disturbances, offering a durable and cost-effective route toward general-purpose fine-grained manipulation in unstructured environments. Project videos and supplementary materials are available at: https://sites.google.com/view/vista-policy.

\end{abstract}

\def\abstractname{Note to Practitioners}
\begin{abstract}
This work focuses on the practical challenge of reliable robot manipulation in contact-rich tasks, where conventional visual observations provide ambiguous contact states, and dedicated tactile hardware remains costly and fragile. We propose a low-cost alternative in which an external camera observes the visible 3D deformation of a passive compliant gripper, and the observed deformation is used directly for policy control. This enables fine-grained manipulation---such as adaptive grasping, cap handling, and tool use--—without dedicated tactile or force sensors. For practitioners, this setup provides a durable, low-maintenance hardware option, high sample efficiency, and strong robustness to object-scale changes, disturbances, and fragile objects. The main practical requirement is that the manipulated system use a compliant gripper whose deformation remains observable from an external camera.
\end{abstract}
\def\abstractname{Abstract}


\section{Introduction}

\begin{figure}[t] 

    \centering

    \includegraphics[width=\linewidth]{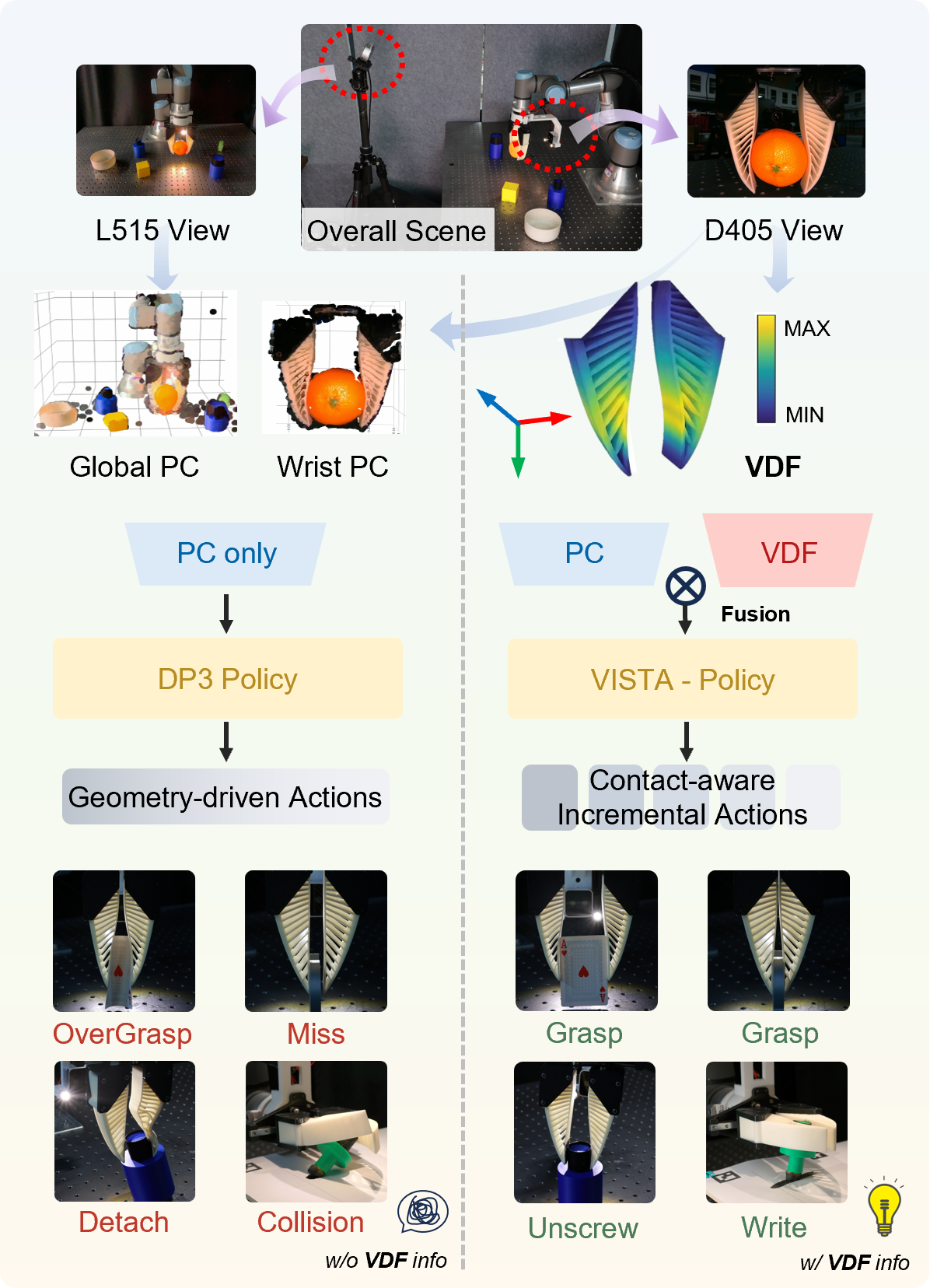} 

    \caption{Overview of VISTA-Policy. By extracting the Visual Deformation Field from a passive compliant gripper, our framework provides high-dimensional visuo-physical feedback for robust contact-rich manipulation without dedicated tactile or force sensors.}

    \label{fig:fig1}
\end{figure}

\IEEEPARstart{C}{ontact-rich} manipulation---such as surface finishing~\cite{liu2025forcemimic}, handling fragile objects~\cite{huang20253d}, and assembly~\cite{heo2025furniturebench, zhu2025shapeforce}---requires robots to possess exceptional dynamic adaptability and compliant interaction~\cite{suomalainen2022survey}. These capabilities are essential to handle uncertainties like manufacturing tolerances and environmental disturbances~\cite{cui2021toward}. Contemporary manipulation policies predominantly rely on visual inputs~\cite{chi2025diffusion, ze20243d, zhao2023learning}, driven by deep learning breakthroughs and low-cost cameras~\cite{dosovitskiy2020image, liu2025fusion}. While these vision-centric policies excel at complex tasks like organizing kitchenware~\cite{black2024pi_0}, they encounter limitations in tasks demanding delicate physical interaction. Importantly, this limitation does not arise from vision itself, but from what conventional visual policies observe: object appearance, geometry and robot motion provide only indirect and ambiguous evidence of contact states, especially when contact regions are occluded or when interaction-induced motions are subtle~\cite{chen2025implicitrdp}. Consequently, the learned policies can be vulnerable to variations in object compliance, geometry, and external disturbances and may require more demonstrations to learn contact-sensitive behaviors~\cite{lee2020making, emukpere2026robust, sferrazza2024power, zhao2025touch, hogan2020tactile}. In contrast, the deformation of a compliant gripper is a visually observable physical response directly induced by contact. Unlike conventional scene-level appearance, such deformation provides a more explicit indication of the underlying interaction state. This observation suggests an alternative route for obtaining contact-relevant feedback through vision.

Dedicated tactile~\cite{huang20253d, feng2025play, li2023see} and force/torque sensors~\cite{zhou2025admittance, chen2023visuo} provide such direct interaction feedback and significantly enhance robotic manipulation capabilities. Nevertheless, these hardware solutions face notable practicality challenges. First, durability constraints and deployment costs. Mainstream visuo-tactile sensors, such as GelSight and others~\cite{yuan2017gelsight, ward2018tactip, lambeta2020digit}, often rely on soft elastomers susceptible to wear~\cite{george2025vital, dong2017improved, davis2025benchmarking}. Similarly, commercial force/torque sensors can be cost-prohibitive or bulky for general-purpose use~\cite{zhu2025shapeforce, choi2026wild, zhu2025forces}. These limit their suitability for long-term, high-intensity operations in unconstrained environments. Second, perception gaps. While highly sensitive, wrist-mounted force/torque sensors struggle to capture finger-level forces, whereas visuo-tactile sensors may encounter difficulties resolving large planar contacts~\cite{yuan2017gelsight, li20233}. Finally, data heterogeneity and cross-platform generalization. Various tactile sensors can exhibit cross-device data inconsistency~\cite{saito2024latent, zhao2025transferable}. Furthermore, processing high-dimensional tactile observations introduces substantial computational overhead, posing challenges for deployment on low-cost platforms. While dedicated sensors indicate the value of contact feedback~\cite{billard2019trends}, the above issues limit large-scale deployment in general-purpose robotic systems. This raises a critical question: Can we extract physically meaningful contact feedback for robotic policies without relying on expensive, fragile hardware?

Biological cross-modal compensation provides a useful source of inspiration. Neuroscience shows that humans can perform complex tasks---like tying shoelaces or writing---even with severe tactile and proprioceptive deficits, relying on visual monitoring of object and hand deformation~\cite{cole1992perceptions}. Inspired by this observation, we investigate whether a robot can similarly exploit visible structural deformation as a proxy for contact interaction. Accordingly, we propose \textbf{VISTA-Policy}, a novel imitation learning (IL) policy within our \textbf{V}isually \textbf{I}nferred \textbf{S}patial Con\textbf{T}act \textbf{A}ttention framework. As illustrated in Fig.~\ref{fig:fig1}, VISTA-Policy relies on a 3D displacement representation, termed the \textbf{Visual Deformation Field (VDF)}, that encodes the deformation of a passive compliant gripper. By extracting the VDF from external visual observations, VISTA-Policy derives structured visuo-physical feedback for downstream manipulation. Its core architecture consists of three key components:

\textit{Physics-Aware Encoding Engine}: It acts as an implicit visual attention mechanism, extracting the camera-view-consistent 3D deformation representation from the observed compliant gripper, transforming appearance changes into physically interpretable interaction cues.

\textit{Energy Aggregation Denoising Mechanism}: A filtering module that calculates deformation energy distributions to isolate valid physical contact signals from environmental noise, suppressing modal interference during non-contact phases.

\textit{Deformation-Augmented Policy Network}: A multimodal architecture that dynamically fuses the deformation modality, enhancing spatio-temporal feature extraction and guiding the policy to map deformation feedback to manipulation actions.

We conduct comprehensive evaluations of both the perception engine and the overall policy. The perception engine achieves accurate deformation extraction and contact identification even under severe visual occlusion. At the policy level, we evaluate VISTA-Policy on three challenging tasks: \textit{Cross-Scale Object Grasping}, \textit{Cap Unscrewing}, and \textit{Calligraphy Writing}. Results show that VISTA-Policy substantially outperforms the strong pure-vision baseline, 3D Diffusion Policy (DP3), and a tactile baseline (equipped with dedicated visuo-tactile sensors) in success rate, data efficiency, and generalization. Furthermore, it exhibits substantial robustness and self-recovery when handling fragile objects, such as soft tofu, or when facing dynamic disturbances. The main contributions of this paper are summarized as follows:

\textbf{First}, we introduce a contact-aware IL paradigm that translates \textbf{observable gripper deformation into visuo-physical feedback}, addressing the limitations of pure-vision approaches without dedicated tactile hardware.

\textbf{Second}, we develop the \textbf{VISTA-Policy} framework for implicit deformation perception and modal fusion. This approach effectively integrates 3D VDF extraction, energy aggregation denoising, and a deformation-augmented policy architecture. 

\textbf{Third}, extensive experiments on contact-rich tasks demonstrate VISTA-Policy's \textbf{superior success rate, cross-scale adaptability, and robustness} over the strong vision baseline DP3 and the tactile baseline.

\begin{figure}[htbp] %

    \centering

    \includegraphics[width=\linewidth]{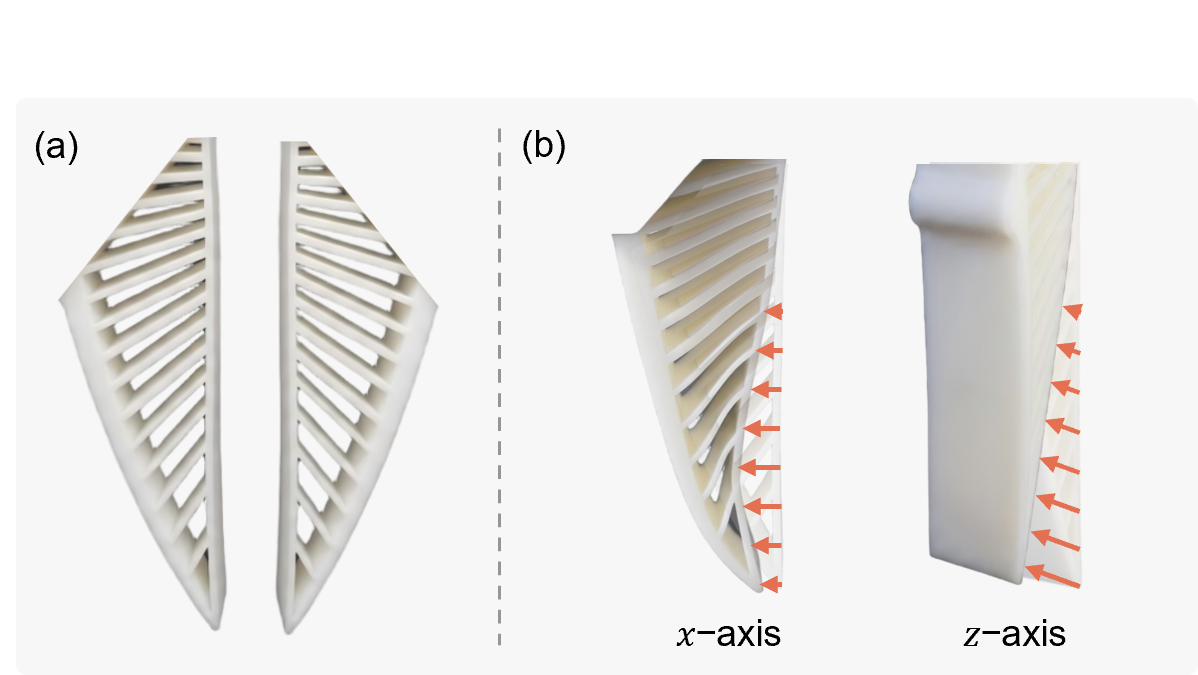} 

    \caption{Structure of the passive compliant gripper and its VDF. (a) Front view of the gripper. (b) Observable reactive deformation along the $x$-axis and $z$-axis under external contact, providing the physical basis for VDF representation.}

    \label{fig:VDF} %

\end{figure}

\section{Related Work}

\subsection{Learning-Based Policies for Robotic Manipulation}

Data-driven methods have emerged as a prevailing paradigm in robotic manipulation. 
Unlike Reinforcement Learning (RL), which often suffers from severe sample inefficiency and sim-to-real transfer gaps, IL receives widespread attention as an efficient approach to transferring expert experience. Although early IL schemes struggled with compounding errors and non-deterministic behaviors, recent breakthroughs in generative architectures significantly mitigate these issues. 
For instance, the Action Chunking Transformer~\cite{zhao2023learning} stabilizes action sequence prediction via temporal ensembling, while Diffusion Policy~\cite{chi2025diffusion} effectively models complex behavioral distributions through iterative denoising. 
Building upon this, DP3~\cite{ze20243d} effectively integrates 3D point clouds to capture scene geometry, exhibiting superior sample efficiency and generalizability. Despite these advancements in trajectory modeling, the execution precision of such policies remains fundamentally constrained by the representational capacity of input observations. For example, point clouds effectively describe object shape and spatial relationships, but do not directly expose contact-induced physical responses, particularly when interaction changes are subtle or occluded. Consequently, relying purely on scene-level observations imposes a critical perceptual bottleneck for fine-grained manipulation requiring precise contact feedback.

Early IL frameworks typically modeled gripper actions as binary states~\cite{rahmatizadeh2018vision}, but this coarse representation is inadequate for fine-grained manipulation. Recent studies, such as FARM~\cite{helmut2025tactile}, explore continuous apertures combined with explicit target force modeling, significantly improving interaction quality. However, these approaches still rely heavily on absolute state representations. Since IL fundamentally models the conditional action distribution bounded by the expert demonstrations, the trained policy struggles to generate out-of-distribution (OOD) actions when encountering unobserved states.
This representational limitation bottlenecks cross-scale generalization and induces spatial overfitting---where the network tends to overfit to specific absolute apertures from the dataset rather than understanding the underlying physical causality of a successful grasp.

To resolve these coupled perception and control issues, our proposed VISTA-Policy integrates the VDF as closed-loop visuo-physical feedback and reformulates the gripper action space into a \textit{relative increment} representation (i.e., the positional delta between consecutive timesteps), while retaining the original arm action representation. This design unbinds the policy from absolute geometric anchors, demanding the network to dynamically modulate its actions guided by real-time visuo-physical feedback.

\subsection{Sensing and Feedback for Contact-Rich Manipulation}

While mainstream vision-centric policies~\cite{dosovitskiy2020image, liu2025fusion, rahmatizadeh2018vision} excel globally, their distal nature struggles to capture high-frequency contact dynamics. Consequently, researchers explore multimodal feedback (e.g., tactile, proprioceptive, and audio modalities) to enhance system robustness. However, despite its theoretical value, tactile sensing currently faces practicality problems. Visuo-tactile sensors~\cite{yuan2017gelsight, ward2018tactip} encounter difficulties like inherent fragility~\cite{george2025vital, dong2017improved, donlon2018gelslim, davis2025benchmarking}, high computational overhead~\cite{dosovitskiy2020image, pattabiraman2024learning}, and measurement failures during planar contacts~\cite{yuan2017gelsight, fang2025force, li20233}. Additionally, wrist-mounted force/torque sensors fail to capture finger-level forces due to internal force closure~\cite{choi2026wild}, while inertial and vibrational noise during non-contact states can heavily disrupt policy learning~\cite{feng2025play}. Moreover, their low-dimensional (6-DoF) data restrict the spatial decoupling of high-frequency noise; thus, temporal filtering often introduces phase delays, which can lead to perception lag and over-execution.

To circumvent these limitations, implicit vision-based force sensing~\cite{zhu2025forces} extracts forces from the deformation of specifically designed articulated grippers via external cameras. While eliminating fragile internal sensors, their heavy reliance on specialized mechanisms and complex kinematics restricts generalizability. Crucially, integrating this implicit representation into data-driven frameworks for robust manipulation remains a challenge.

Our proposed VDF intrinsically addresses these limitations. Directly extracted from an external camera, VDF naturally aligns with visual states, mitigating complex cross-modal registration. Crucially, its dense sampling yields high spatial resolution. While minor vibration and tracking noise typically induce random low-amplitude perturbations, genuine contact produces coherent deformation across neighboring regions of the compliant fingers. This empowers the system to spatially decouple noise from interaction signals in real-time, providing reliable contact-state feedback for downstream policy learning.

\subsection{OOD Generalization and Physical Invariants}

While multimodal sensing facilitates fine-grained manipulation, achieving OOD generalization across unseen categories or drastic scale variations remains a critical challenge for IL policies. Current solutions predominantly follow two paths. 

First, large-scale Vision-Language-Action (VLA) models (e.g., $\pi_0$~\cite{black2024pi_0}) exhibit strong semantic generalization. However, their high inference latency and lack of low-level physical precision (e.g., torque modulation) render them suboptimal for high-frequency closed-loop control. 

Second, geometric inductive biases (e.g., $SE(3)$-equivariant networks) are widely used to extract ``task invariants". Notably, frameworks like VITAL~\cite{zhao2025touch} demonstrate the high consistency of local physical interactions by decoupling global reaching from local manipulation. Nevertheless, these methods typically necessitate specialized tactile hardware (e.g., AnySkin~\cite{bhirangi2025anyskin}) or extensive online fine-tuning via residual RL.

This study posits that achieving OOD generalization requires a representation that bypasses complex hardware constraints while accurately capturing the underlying physical essence. The VDF of the compliant gripper serves as an explicit, low-cost physical invariant. Because reactive deformations strictly obey underlying mechanics, they exhibit intrinsic physical consistency across grasping tasks involving diverse geometries and materials. By injecting this lightweight visuo-physical prior into data-driven frameworks like DP3, we aim to equip IL policies with robust cross-category and cross-scale grasping capabilities without adding external sensory payloads, effectively bridging the gap between pure geometric observation and complex tactile perception.

\section{Method}

\subsection{Overview}
\begin{figure*}[t] 

    \centering

    \includegraphics[width=\linewidth]{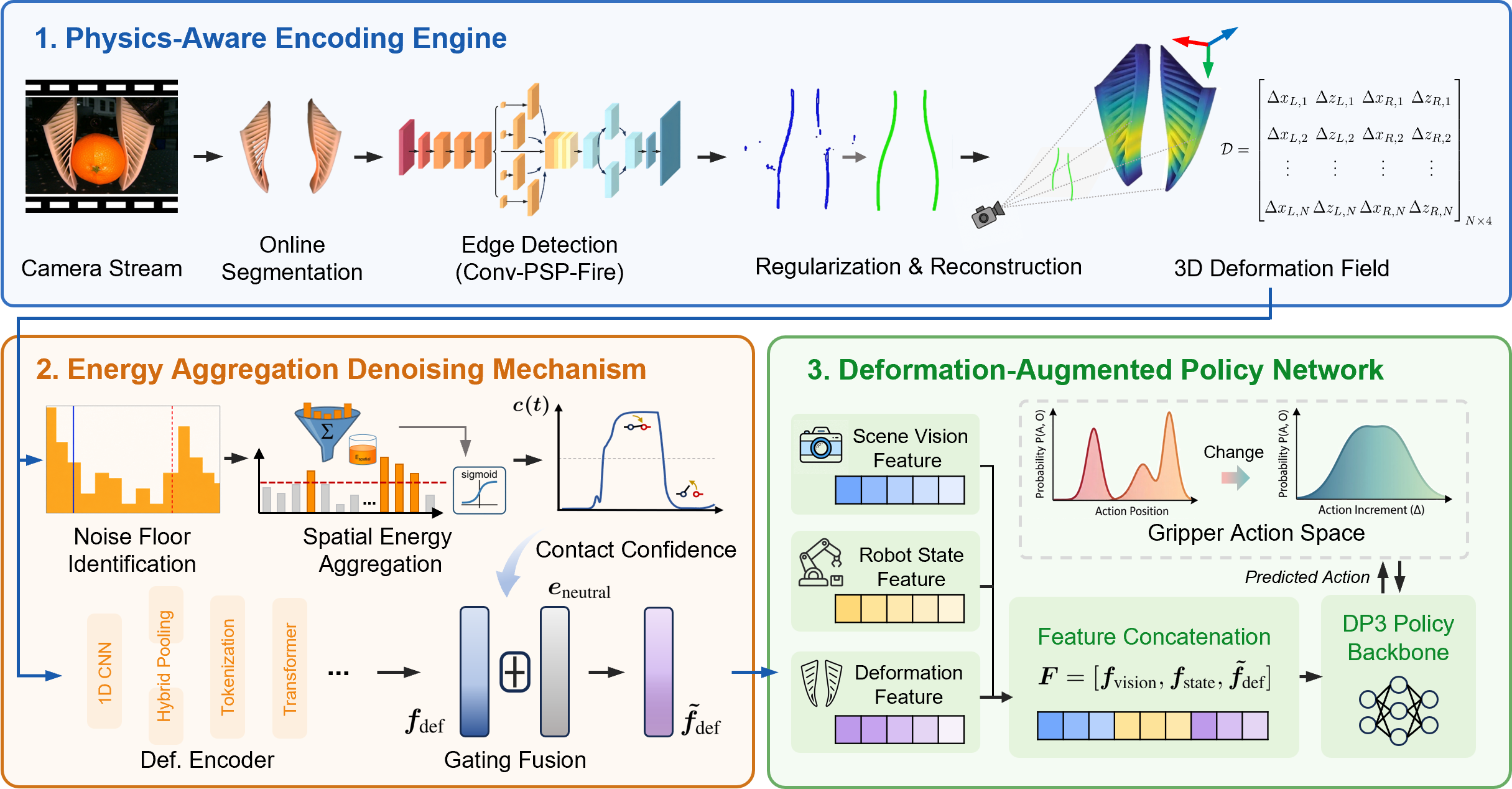} 

    \caption{Architecture of VISTA-Policy, comprising three core components: a \textit{Physics-Aware Encoding Engine} for 3D VDF extraction, an \textit{Energy Aggregation Denoising Mechanism} for contact denoising, and a \textit{Deformation-Augmented Policy Network} driven by a relative incremental gripper action space.}

    \label{fig3-diagram} 

\end{figure*}

As illustrated in Fig.~\ref{fig3-diagram}, VISTA-Policy comprises three core stages: 1) The \textit{Physics-Aware Encoding Engine} (Section~\ref{sec:perception}) extracts a structured 3D VDF $\bm{D}$ from real-time camera inputs $I_t$. 2) The \textit{Energy Aggregation Denoising Mechanism} (Section~\ref{sec:energy}) extracts a continuous contact confidence $c_t$ from $\bm{D}$, guiding a gating mechanism to output a fused global representation $\bm{F}$. 3) The \textit{Deformation-Augmented Policy Network} (Section~\ref{sec:policy}) leverages a spatio-temporal encoder and a relative incremental gripper action space to generate robust manipulation commands.

\subsection{Physics-Aware Encoding Engine}
\label{sec:perception}

To achieve reliable 2D segmentation and tracking of the compliant gripper, we build upon Track-Anything~\cite{yang2023track} as our foundational visual tracker. However, its native batch-processing paradigm incurs an $\mathcal{O}(T)$ memory complexity with temporal sequence length $T$, which violates the real-time constraints of closed-loop robotic manipulation. To overcome this, we reconstruct the architecture into a state-machine-driven online streaming pipeline. At time $t$, the tracker $\Phi$ executes single-step inference using only the current frame $I_t$ and an implicit memory feature $\mathcal{M}_{t-1}$:
\begin{equation}
(M_{t}, \mathcal{M}_{t}) = \Phi(I_{t}, \mathcal{M}_{t-1}),
\end{equation}
where $M_t$ denotes the output binary mask, and $\mathcal{M}_t$ represents the updated implicit memory feature for the next iteration. This temporal decoupling strictly bounds the complexity to $\mathcal{O}(1)$, mitigating memory overflow risks. Supported by asynchronous computation optimizations, the pipeline reliably operates at 20 Hz with minimal impact on precision, thereby meeting high-frequency closed-loop control requirements.

To handle real-world occlusions and morphological variations, the algorithm focuses feature extraction on the anterior inner edge line of the gripper, which offers high occlusion resilience and directly reflects deformation. To this end, we design a lightweight multi-scale semantic segmentation network: the front end utilizes large convolutional kernels and residual modules for rapid downsampling while ensuring robust gradient propagation under extreme illumination; a pyramid scene parsing module then fuses multi-scale contexts to mend occlusion-induced feature fragmentation; finally, a Fire module compresses feature dimensions to minimize computational overhead.

To rectify artifacts and fractures in the segmented masks, a rule-based post-processing module is introduced. First, morphological opening and area-threshold filtering extract clean edge mask clusters. Subsequently, the left and right edge point sets are independently parameterized using B-spline curves. This step reconstructs discrete, incomplete masks into smooth, continuous geometric profiles, establishing a high-precision geometric baseline for subsequent 3D VDF estimation.

Utilizing aligned RGB-D depth maps, the system back-projects 2D edge pixels into 3D space to reconstruct the VDF. We initialize a local 3D coordinate system anchored at the gripper's rigid root and perform equidistant sampling at $N$ points along the longitudinal $y$-axis. The effect of the VDF is illustrated in Fig.~\ref{fig:VDF}. At each sampling index $i$, the structural deformation of both the left and right finger edges is constrained within the $xz$-plane and formalized as a combined relative displacement vector at timestep $t$:
\begin{equation}
\bm{d}_{i,t} = [\Delta x_L, \Delta z_L, \Delta x_R, \Delta z_R] \in \mathbb{R}^4,
\end{equation}
where subscripts $L$ and $R$ denote the left and right gripper fingers, respectively. By strictly defining deformations as dynamic differences relative to the rigid root, we eliminate spurious correlations between absolute gripper positions and contact states, ensuring physical invariant feature representations. The $y$-coordinates are implicitly encoded via their sampling indices and thus omitted. Finally, to handle mask fractures caused by adverse lighting or occlusions, a depth completion module executes an independent 1D linear interpolation across each feature column, filling missing values using adjacent valid $z$ measurements.

The final output is a structured 3D VDF represented as $\bm{D} \in \mathbb{R}^{T \times N \times C}$, where $T$ denotes the observation horizon, $N$ is the aforementioned spatial sampling dimension, and $C=4$ corresponds to the displacement feature channels in $\bm{d}_{i,t}$. This spatially continuous and fragmentation-free deformation field provides a high-quality data representation that ensures input consistency for the subsequent policy network.

\subsection{Energy Aggregation Denoising Mechanism}
\label{sec:energy}
Directly feeding the raw VDF into the network renders the model susceptible to background noise during non-contact phases. Rather than relying solely on temporal filtering, the spatial structure of the VDF allows us to distinguish coherent contact-induced deformation from minor vibrations and isolated tracking noise. Hence, we design a denoising scheme based on spatial energy aggregation. This mechanism filters out global noise in real-time, transforming the high-dimensional, noisy VDF into a compact, physically meaningful contact confidence $c_t$, which guides the policy network to incorporate the deformation modality at the appropriate juncture.

To eliminate minor perturbations during non-contact states, we first calibrate a deformation noise floor threshold $\tau_{\text{noise}}$. Because isolated visual tracking outliers can easily exceed the threshold $\tau_{\text{noise}}$ and trigger false positives, relying on the deformation magnitude of a single point for contact determination is insufficiently robust. Therefore, after filtering out the noise floor, we integrate the effective deformation vectors across the global field to accurately determine the contact state. We employ the ReLU activation function as a deadzone filter to truncate invalid signals below $\tau_{\text{noise}}$, computing the effective spatial energy $E_{\text{spatial}, t}$ at time $t$:
\begin{equation}
E_{\text{spatial}, t} = \frac{1}{N} \sum_{i=1}^{N} \text{ReLU}(\|\bm{d}_{i, t}\|_2 - \tau_{\text{noise}}),
\end{equation}
where $N$ is the total number of spatial sampling points, and $\|\bm{d}_{i, t}\|_2$ denotes the deformation magnitude of the $i$-th point.

With a calibrated nominal trigger center $E_{\text{mid}}$ established for early contact detection, we utilize a sigmoid function to map the physical energy scalar into a normalized contact confidence:
\begin{equation}
c_{\text{raw}, t} = \frac{1}{1 + e^{-k(E_{\text{spatial}, t} - E_{\text{mid}})}},
\end{equation}
where the gain coefficient $k$ controls the signal trigger sensitivity, enabling $c_{\text{raw}, t}$ to transition sharply from 0 to 1 when the average deformation exceeds the critical threshold, thereby establishing a distinct state-switching feature. To prevent transient jumps caused by momentary visual occlusions or dynamic motion blur, an exponential moving average (EMA) is applied to obtain a temporally continuous contact confidence $c_t$:
\begin{equation}
c_t = \alpha \cdot c_{\text{raw}, t} + (1 - \alpha) \cdot c_{t-1},
\end{equation}
where $\alpha$ is the smoothing factor. 

Finally, to resolve the interference introduced by deformation data during non-contact phases, we introduce a gating fusion mechanism that utilizes $c_t$ to perform a soft gating operation on the deformation features. Specifically, we define a learnable neutral embedding vector $\bm{e}_{\text{neutral}}$ to represent the nominal non-contact state. The gated feature $\bm{\tilde{f}}_{\text{def}}$ routed to the downstream network is computed as:
\begin{equation}
\bm{\tilde{f}}_{\text{def}} = c_t \cdot \bm{f}_{\text{def}} + (1 - c_t) \cdot \bm{e}_{\text{neutral}}.
\end{equation}
Ultimately, the gated deformation features, the scene visual features $\bm{f}_{\text{vision}}$, and the robot proprioceptive state $\bm{f}_{\text{state}}$ are concatenated to form a comprehensive multimodal state representation $\bm{F}$ for the downstream policy network:
\begin{equation}
\bm{F} = [\bm{f}_{\text{vision}}, \bm{f}_{\text{state}}, \bm{\tilde{f}}_{\text{def}}].
\end{equation}

\subsection{Deformation-Augmented Policy Network}
\label{sec:policy}
Traditional IL frameworks predominantly employ absolute positions to represent actions. This not only restricts the network's ability to generate OOD actions but also hinders cross-scale generalization. To address this, we reconstruct the gripper action space into a relative increment representation. During deployment, the incremental action predicted by the policy is superposed onto the current actual opening. This design unbinds the policy from absolute action space constraints, enabling the model to dynamically adjust the gripper closure guided by the real-time deformation states.

To empower the model to accurately interpret and exploit the contact-rich information embedded within our VDF modality, we design a spatio-temporal deformation encoder. First, a 1D CNN processes the single-frame VDF. This design treats the $N$ sampling points as a sequence with inherent physical topology, thereby effectively learning localized deformation patterns. To enhance the model's robustness across varying contact areas, we introduce a dual-pooling strategy: adaptive max pooling is used to capture peak deformation, while adaptive average pooling encodes the overall contact distribution. The fused spatial embedding sequence $\{\bm{f}_1, \bm{f}_2, \dots, \bm{f}_T\}$ is subsequently fed into a temporal Transformer network. We introduce a learnable readout token $\bm{x}_{\text{readout}}$ to serve as a global proxy for the entire sequence, leveraging a multi-head self-attention mechanism to aggregate the spatio-temporal contact dynamics of the interaction task.
\begin{equation}
\bm{X}_{\text{in}} =
[\bm{x}_{\text{readout}}, \bm{f}_1 + \bm{p}_1, \dots, \bm{f}_T + \bm{p}_T],
\end{equation}
\begin{equation}
\bm{X}_{\text{out}} =
\text{TransformerEncoder}(\bm{X}_{\text{in}}).
\end{equation}
The final readout token feature serves as the global contact-state representation for the manipulation window, which is then aligned with the downstream policy network.

\section{Experimental Setup}

Our hardware platform comprises a UR3 robotic arm, a UMI-based passive compliant gripper~\cite{chi2024universal}, and RealSense L515 and D405 depth cameras for global and local observations, respectively. To comprehensively evaluate our method, we systematically design the following comparative groups: \textbf{1) Primary Baselines:} \textbf{DP3}: Standard DP3 utilizing only the global point cloud (PC). \textbf{DP3-Wrist}: DP3 augmented with local wrist-view PC (ensuring viewpoint fairness). \textbf{TDF-DM}: A hardware tactile baseline replacing the compliant gripper with a commercial Daimon visuo-tactile sensor. \textbf{VISTA-Policy (ours)}: Our full pipeline that combines VDF feedback with a relative incremental gripper action space. \textbf{2) Ablation Baselines:} To decouple the contribution of specific modules, we introduce the following variants. \textbf{Action Space Ablations (\textit{-Abs})}: \textbf{VISTA-Abs} and \textbf{DP3-Wrist-Abs} modify the respective base models to output absolute gripper action spaces rather than relative increments, isolating the impact of the action representation. \textbf{Architecture Ablations}: \textbf{VISTA-Cat} concatenates deformation features without energy denoising to evaluate the contribution of energy-based denoising. \textbf{VISTA-MLP} replaces the spatio-temporal encoder with an MLP to evaluate our structured deformation processing. \textbf{3) Sample Size Variants:} Unless otherwise specified, all models are trained on 20 expert demonstrations. For sample efficiency evaluations, \textbf{DP3-Wrist-40} denotes DP3-Wrist trained with 40 demonstrations. A detailed summary of the components for all methods is presented in Table~\ref{table:baseline_components}.

\begin{table}[htbp]
\centering
\caption{Systematic Comparison of Component Configurations between Different Methods}
\label{table:baseline_components}
\renewcommand{\arraystretch}{1.15}
\setlength{\tabcolsep}{3pt} 
\begin{tabular}{lccccc}
\toprule
\multirow{2}{*}{\textbf{Method}} & \multicolumn{4}{c}{\textbf{Observation Modality}} & \multirow{2}{*}{\begin{tabular}[c]{@{}c@{}}\textbf{Rel. Gripper}\\\textbf{Action ($\Delta$)}\end{tabular}} \\
\cmidrule(lr){2-5}
& Global PC & Wrist PC & VDF & Tactile & \\
\midrule
DP3-Wrist-Abs    & \cmark & \cmark & \xmark & \xmark & \xmark \\
VISTA-Abs        & \cmark & \xmark & \cmark & \xmark & \xmark \\
\midrule
DP3              & \cmark & \xmark & \xmark & \xmark & \cmark \\
DP3-Wrist        & \cmark & \cmark & \xmark & \xmark & \cmark \\
TDF-DM           & \cmark & \xmark & \xmark & \cmark & \cmark \\
VISTA-Cat        & \cmark & \xmark & \cmark & \xmark & \cmark \\
VISTA-MLP        & \cmark & \xmark & \cmark & \xmark & \cmark \\
\midrule
\textbf{VISTA (\textit{ours})} & \textbf{\cmark} & \xmark & \textbf{\cmark} & \xmark & \textbf{\cmark} \\
\bottomrule

\end{tabular}
\end{table}

\begin{figure}[t] 
    \centering
    \includegraphics[width=\linewidth]{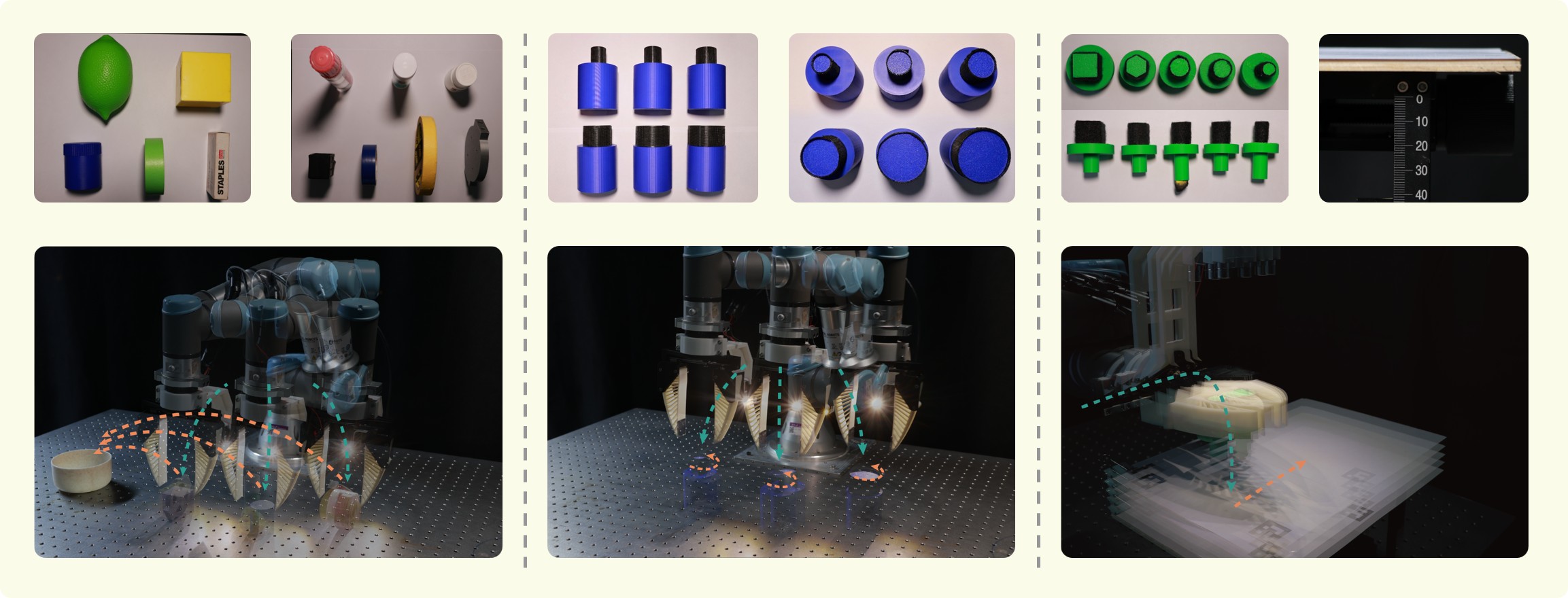} 
    \caption{Overview of the three contact-rich manipulation tasks: (left) \textit{Cross-Scale Object Grasping} of diverse objects; (center) \textit{Cap Unscrewing} with varied diameters; (right) \textit{Calligraphy Writing} on platforms of different heights. The top row illustrates the object sets, while the bottom row shows the task execution sequences.}
    \label{fig:task_overview}
\end{figure}

We design three contact-rich manipulation tasks with progressive difficulty gradients (Fig.~\ref{fig:task_overview}): \textit{Cross-Scale Object Grasping}: The robot must grasp and transport objects of different widths, requiring the policy to adaptively modulate gripping closure across scales, testing deformation perception in the $xz$-plane. \textit{Cap Unscrewing}: The robot must localize and unscrew bottle caps of varying diameters, demanding perception of depth ($z$-axis) deformation. \textit{Calligraphy Writing}: The robot manipulates brushes of varying geometries and diameters to write horizontal strokes on platforms of different heights, requiring fluid motion while maintaining stable $z$-axis deformation. Across these tasks, we evaluate cross-scale generalization, action-space ablation, sample efficiency and robustness. Standard task evaluations use 10 independent trials per configuration. Additional robustness evaluations, including dynamic disturbances, fragile-object manipulation, and extreme OOD settings, use 5 trials per configuration.

We quantify performance from three perspectives: task completion, execution quality, and physical consistency. The primary metric is Success Rate (SR), defined as canonical full-task completion. We further report fine-grained quality metrics, including Fall-off Rate (FR) for grasping, Dislocation Rate (DR) for \textit{Cap Unscrewing}, and Holding/Writing Success Rates (HSR/WSR) for \textit{Calligraphy Writing}. For writing quality, we use the largest connected region ratio ($LCR$) to measure stroke continuity and $Solidity$ to measure stroke fullness. We then define the Stroke Quality Index ($SQI$) as:
\begin{equation}
SQI = \sqrt{LCR \times {Solidity}}.
\end{equation}
 Crucially, we analyze numerical statistics of the VDF to verify whether the policy learns visuo-physical closed-loop control rather than mere visual servoing.

\begin{figure*}[!t] 

    \centering

    \includegraphics[width=\textwidth]{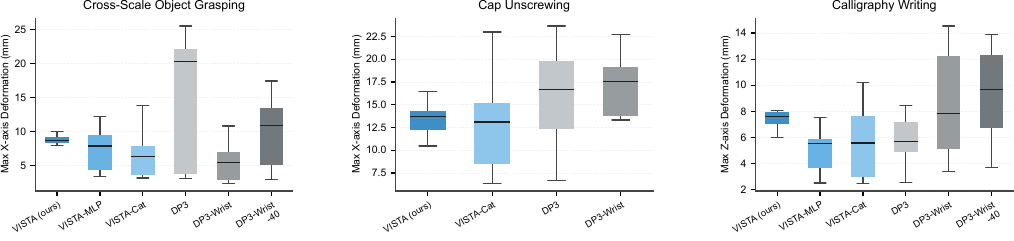} 

    \caption{Statistical analysis of VDF during the contact phase across the three tasks. VISTA-Policy exhibits lower dispersion than the compared baselines, indicating more consistent contact-induced deformation during execution. (For methods that do not use VDF as policy input, VDF is computed only for post-hoc evaluation.)}

    \label{fig:11} 

\end{figure*}%
\vspace{-2pt}

\section{Experiments}
Through a series of systematic evaluation trials, this research aims to address the following three core research questions:
\textbf{Q1:} Does the proposed VDF modality provide a more consistent and robust physical contact representation compared to raw visual modalities?
\textbf{Q2:} Can VISTA-Policy substantially enhance execution performance and OOD generalization in contact-rich manipulation tasks?
\textbf{Q3:} Does VISTA-Policy enable superior robustness, higher sample efficiency and safer physical interaction?

Prior to policy evaluation, we conduct extensive evaluations on the \textit{Physics-Aware Encoding Engine} to verify the robustness of edge extraction. Experimental details are provided in the Supplementary Material.

\subsection{General Analysis}

\begin{figure}[!t] 

    \centering

    \includegraphics[width=\columnwidth]{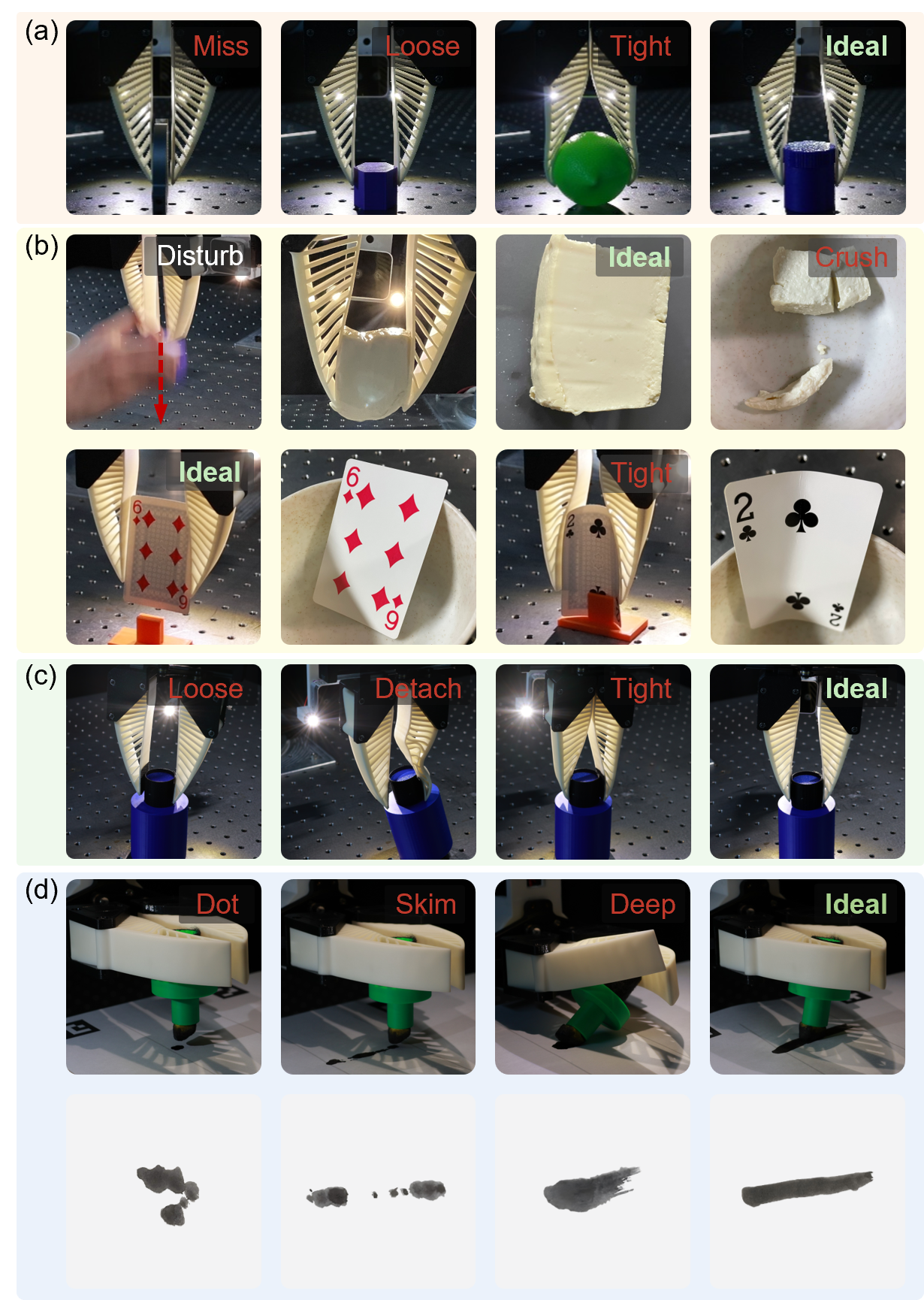} 

    \caption{
    Demonstration of various execution situations across the three tasks. (a) Several grasping situations in the \textit{Grasping} task. (b) Robustness evaluations in the \textit{Grasping} task, including dynamic disturbances and the manipulation of tofu and playing cards. (c) Different execution situations in the \textit{Cap Unscrewing} task. (d) Various execution situations in the \textit{Calligraphy Writing} task and the corresponding stroke effects.
    }

    \label{qualitative_overview} 

\end{figure}

Qualitative execution scenarios across three tasks are illustrated in Fig.~\ref{qualitative_overview}. As shown in Table~\ref{table:task_evaluation_v2}, VISTA-Policy comprehensively outperforms all baselines across all evaluated aspects of the three contact-rich tasks. This significant enhancement in performance supports \textbf{Q2}.

Beyond macro-level success rates, Fig.~\ref{fig:11} illustrates the gripper deformation characteristics during the contact phases. VISTA-Policy consistently exhibits the narrowest fluctuation range across all tasks compared to the baselines. This substantial physical consistency confirms that VISTA-Policy accurately captures contact regularities based on the VDF, thereby generating \textbf{appropriate and reliable control commands}, effectively answering \textbf{Q1}.

\begin{table}[t]
\centering
\caption{Evaluation Results of the Three Tasks}
\label{table:task_evaluation_v2}
\renewcommand{\arraystretch}{1.15}
\footnotesize 
\setlength{\tabcolsep}{1.2pt} 
\begin{tabular}{@{}lcccccc@{}}
\toprule
\multirow{2}{*}{\textbf{Method}} & \multicolumn{2}{c}{\textbf{\makecell{Cross-Scale\\Object Grasping}}} & \multicolumn{2}{c}{\textbf{\makecell{Cap\\Unscrewing}}} & \multicolumn{2}{c}{\textbf{\makecell{Calligraphy\\Writing}}} \\
\cmidrule(lr){2-3} \cmidrule(lr){4-5} \cmidrule(lr){6-7}
& SR (\%)$\uparrow$ & FR (\%)$\downarrow$ & SR (\%)$\uparrow$ & DR (\%)$\downarrow$ & HSR (\%)$\uparrow$ & WSR (\%)$\uparrow$ \\
\midrule
DP3            & 40  & 70 & 50 & 20 & 50  & 40 \\
DP3-Wrist      & 50  & 50 & 40 & 30 & 90  & 40 \\
TDF-DM         & 60  & 60 & 50 & 80 & 80  & 20 \\
\midrule
\textbf{VISTA (\textit{ours})} & \textbf{100} & \textbf{0}  & \textbf{90} & \textbf{0}  & \textbf{100} & \textbf{90} \\
\bottomrule
\end{tabular}
\end{table}

\begin{table}[t] %
\centering
\caption{OOD Scale Generalization Results and Gripper Action Space Ablation}
\label{table:single_comparison}
\renewcommand{\arraystretch}{1.1}
\resizebox{\columnwidth}{!}{ 
\begin{tabular}{lcccc}
\toprule
\multirow{2}{*}{\textbf{Method}} & \multicolumn{2}{c}{Seen (4\,cm)} & \multicolumn{2}{c}{Unseen} \\
\cmidrule(lr){2-3} \cmidrule(lr){4-5}
& SR (\%) $\uparrow$ & FR (\%) $\downarrow$ & SR (\%) $\uparrow$ & FR (\%) $\downarrow$ \\
\midrule
DP3-Wrist-Abs      & 100 & 0 & 50 & 50 \\
VISTA-Abs      & 100 & 0 & 40 & 60 \\
DP3-Wrist & 60  & 40 & 50 & 50 \\
\midrule
\textbf{VISTA (\textit{ours})} & \textbf{100} & \textbf{0} & \textbf{100} & \textbf{0} \\
\bottomrule
\end{tabular}
} 
\end{table}

\subsection{Cross-Scale Object Grasping}

\subsubsection{Single-Object Training and Cross-Scale Generalization} 
To evaluate OOD cross-scale generalization, the policy is trained using demonstrations collected from a single 4\,cm-wide object. Quantitative results are reported in Table~\ref{table:single_comparison}. VISTA-Policy achieves a substantial performance leap over DP3-Wrist. Empirically, the baseline gripper ceases closure upon reaching the approximate width of the training object (4\,cm) and prematurely enters the lifting phase regardless of contact status, resulting in a 0\% SR for objects narrower than 4\,cm. Conversely, VISTA-Policy adaptively modulates optimal gripper closure across all object scales, achieving a 100\% SR and demonstrating robust extrapolation to objects smaller than the training samples. This substantial cross-scale adaptability resolves \textbf{Q2}, indicating that embedding VDF as a physical invariant empowers the policy to effectively \textbf{capture the underlying mechanics of grasping beyond the demonstrated object scales}.

\subsubsection{Gripper Action Representation Ablation}

As shown in Table~\ref{table:single_comparison}, DP3-Wrist-Abs and VISTA-Abs successfully grasp the 4\,cm training object but generalize poorly to unseen scales. VISTA-Abs severely over-grips the 5\,cm object; intriguingly, when facing narrow targets (1 and 2\,cm), the gripper falls into a stall characterized by repetitive minor opening-and-closing oscillations upon reaching a closure of approximately 4\,cm. In contrast, DP3-Wrist-Abs completely ignores the lack of contact, proceeding to ascend with an empty grasp.

These behavioral differences demonstrate that the VDF modality forces the policy to continuously monitor valid contact states (which inherently induces the stall in VISTA-Abs) and also reveal a modality conflict between the absolute action space and the VDF. Because the minimum action label in the training set corresponds to a 4\,cm width, the absolute policy struggles to breach this numerical boundary to generate smaller apertures during deployment. However, lacking physical contact, the VDF feedback remains zero, deviating substantially from the successful grasp profile learned during training. Consequently, the policy refuses to proceed to the lifting phase, inducing a deadlock stall. Reconfiguring the gripper action space into relative increments fundamentally liberates the system from numerical boundaries, enabling the policy to utilize the VDF as the \textbf{key criterion for adaptive gripper regulation}. Ultimately, the comparative evaluations in Table~\ref{table:single_comparison} rigorously substantiate the necessity of synergizing the VDF modality with the relative incremental gripper action space, resolving \textbf{Q2} while demonstrating the paradigm's strong OOD generalization capability in contact-rich tasks.

\begin{table}[htbp] 
\centering
\caption{Performance Comparison Under Multi-Scale Training Conditions}
\label{table:multi_comparison}
\renewcommand{\arraystretch}{1.1}
\resizebox{\columnwidth}{!}{ 
\begin{tabular}{lcccc}
\toprule
\multirow{2.5}{*}{\textbf{Method}} & \multicolumn{2}{c}{Seen} & \multicolumn{2}{c}{Unseen} \\
\cmidrule(lr){2-3} \cmidrule(lr){4-5}
& SR (\%) $\uparrow$ & FR (\%) $\downarrow$ & SR (\%) $\uparrow$ & FR (\%) $\downarrow$ \\
\midrule
DP3                & 40 & 60 & 40 & 80 \\
DP3-Wrist     & 60 & 40 & 40 & 60 \\
VISTA-Cat          & 80 & 80 & 20 & 80 \\
VISTA-MLP          & 80 & 40 & 40 & 60 \\
TDF-DM           & 80 & 40 & 40 & 60 \\
\midrule
\textbf{VISTA (\textit{ours})} & \textbf{100} & \textbf{0} & \textbf{100} & \textbf{0} \\
\bottomrule
\end{tabular}
}
\end{table}

\subsubsection{Multi-Object Training and Sample Efficiency}
\begin{figure}[t] 

    \centering

    \includegraphics[width=\linewidth]{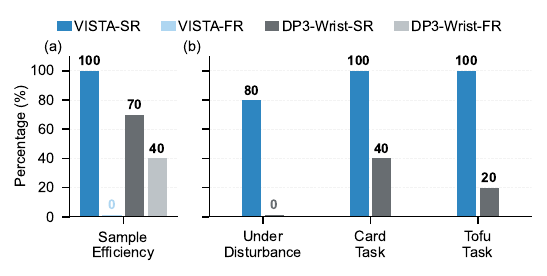} 

    \caption{Quantitative evaluation of sample efficiency and disturbance robustness in Cross-Scale Object Grasping.}

    \label{fig:grasping_metrics} 

\end{figure}

Next, we conduct multi-object grasping experiments to evaluate the efficacy of our paradigm. The expert dataset contains demonstrations across five object diameters: 1.5, 2, 3, 4, and 6\,cm. During deployment, each policy undergoes ten independent trials, comprising one evaluation for each of the five in-distribution training objects and five additional unseen scale-generalization objects. As reported in Table~\ref{table:multi_comparison}, VISTA-Policy achieves 100\% grasping SR across all scales with zero drops. In contrast, the DP3-Wrist baseline struggles to grasp narrow training targets (1.5 and 2\,cm) while exhibiting extreme, excessive gripping (corresponding to deformation exceeding 20\,mm) on larger objects. This gap indicates that despite multi-scale exposure, conventional visual policies fail to master correct manipulation regularities without precise contact-state guidance. Furthermore, although doubling the training samples in DP3-Wrist-40 marginally improves performance over DP3-Wrist, it still induces over-gripping on medium-to-large objects, confirming that the visual baseline cannot distinguish the physical variance between large and small objects. Crucially, the quantitative results in Fig.~\ref{fig:grasping_metrics}(a) reveal that the SR of DP3-Wrist-40 remains inferior to VISTA-Policy trained on only 20 demonstrations. This stark contrast directly answers \textbf{Q3} regarding sample efficiency, demonstrating that the structured VDF guidance provides an \textbf{effective contact cue}, enabling superior performance with limited demonstration data.

\subsubsection{Robustness and Recovery Evaluation}
Qualitative execution behaviors under dynamic disturbances and delicate manipulation scenarios are illustrated in Fig.~\ref{qualitative_overview}(b).
To probe system resilience against environmental disturbances, we apply unexpected dynamic disturbances during execution. Specifically, while the gripper is lifting an object after a successful grasp, the target is manually knocked back onto the tabletop to evaluate whether the closed-loop policy can trigger a secondary recovery grasp. As shown in Fig.~\ref{fig:grasping_metrics}(b), VISTA-Policy completes secondary grasps in 4 out of 5 trials (80\% SR), whereas the DP3-Wrist fails completely. Upon target dislocation, DP3-Wrist exhibits no adaptive response and continues its upward trajectory as if the object were still secured. This outcome validates \textbf{Q3}, confirming that the VDF modality equips the policy with a \textbf{clear understanding of task causality}, ensuring superior robustness against disturbances.

Furthermore, we design delicate manipulation tasks using fresh tofu and playing cards. Fresh tofu represents a highly fragile, heavy target vulnerable to slipping and structural cracking, demanding exceptionally precise closed-loop interaction regulation; its surface also continuously exudes moisture, creating residues that contaminate the interaction interface. Conversely, playing cards are extremely thin and delicate, easily buckling under excessive gripping. The DP3-Wrist either fails to lift the tofu or crushes it entirely due to excessive gripping; when grasping cards, it suffers from drops due to oscillatory jitter or induces visible creases through over-gripping. In sharp contrast, VISTA-Policy successfully manipulates both tofu and cards with adaptive compliance, completing all trials without structural damage. These empirical results firmly resolve \textbf{Q3}, showcasing that the VISTA-Policy architecture possesses real-time perception-action self-recovery, substantial structural robustness, and versatile task universality when confronting unpredictable environmental disturbances and fragile physical boundaries.

\begin{table*}[t] 
\centering
\caption{Generalization Results for Cap Unscrewing under Boundary-Scale Single-Object Training}
\label{table:bottle_capping_scales}
\renewcommand{\arraystretch}{1.1}
\resizebox{0.95\textwidth}{!}{ 
\setlength{\tabcolsep}{8pt} 
\begin{tabular}{lcccccccc}
\toprule
\multirow{3}{*}{\textbf{Method}} & \multicolumn{4}{c}{\textbf{Minimum-Scale Training (2\,cm)}} & \multicolumn{4}{c}{\textbf{Maximum-Scale Training (6\,cm)}} \\
\cmidrule(lr){2-5} \cmidrule(lr){6-9}
& \multicolumn{2}{c}{Seen} & \multicolumn{2}{c}{Unseen} & \multicolumn{2}{c}{Seen} & \multicolumn{2}{c}{Unseen}\\
\cmidrule(lr){2-3} \cmidrule(lr){4-5} \cmidrule(lr){6-7} \cmidrule(lr){8-9}
& SR (\%) $\uparrow$ & DR (\%) $\downarrow$ & SR (\%) $\uparrow$ & DR (\%) $\downarrow$ & SR (\%) $\uparrow$ & DR (\%) $\downarrow$ & SR (\%) $\uparrow$ & DR (\%) $\downarrow$ \\
\midrule
DP3-Wrist & 20 & 40 & 20 & 60 & 60 & 20 & 40 & 20 \\ 
\midrule
\textbf{VISTA (\textit{ours})} & \textbf{100} & \textbf{0} & \textbf{80} & \textbf{20} & \textbf{80} & \textbf{40} & \textbf{80} & \textbf{0} \\
\bottomrule
\end{tabular}
}
\end{table*}%

\subsection{Cap Unscrewing}

\subsubsection{Boundary-Scale Single-Object Training}

To evaluate cross-scale generalization from the two boundary scales, we construct two single-object training settings: a minimum-scale setting trained only on 2\,cm-diameter caps and a maximum-scale setting trained only on 6\,cm-diameter caps. Quantitative results are summarized in Table~\ref{table:bottle_capping_scales}. For the former setting, deployment results indicate that the DP3-Wrist delivers suboptimal performance, successfully unscrewing the cap in only 1 out of 5 trials at the training scale. Empirically, although the baseline policy guides the gripper to a width suitable for the 2\,cm cap, the axis of the end-effector frequently deviates from that of the bottle during rotation, leading to gripper misalignment or severe asymmetric contact that dislocates the bottle. For large-diameter caps (5 and 6\,cm), matching the failure patterns in the grasping task, the baseline generates excessive gripper closure, pulling the bottle off the tabletop almost immediately upon the onset of twisting. In contrast, VISTA-Policy achieves a substantially higher SR with noticeably smoother execution, demonstrating that the VDF provides effective \textbf{contact-state guidance}. In the maximum-scale setting, DP3-Wrist fails entirely to unscrew the small 2 and 3\,cm caps, while VISTA-Policy dynamically regulates the bilateral gripping state based on the VDF, maintaining stable cap engagement during twisting. This asymmetric performance degradation of the baseline under extreme boundary conditions, contrasted with the stable execution of our paradigm, further resolves \textbf{Q2}.

\begin{table}[t] 
\centering
\caption{Quantitative Results on Multi-Object Cap Unscrewing Experiments}
\label{table:multi_bottle_capping}
\renewcommand{\arraystretch}{1.1}
\resizebox{\columnwidth}{!}{ 
\setlength{\tabcolsep}{20pt} 
\begin{tabular}{lcc}
\toprule
\textbf{Method} & \textbf{SR (\%)} $\uparrow$ & \textbf{DR (\%)} $\downarrow$ \\
\midrule
DP3             & 50 & 20 \\
DP3-Wrist  & 40 & 30 \\
VISTA-Cat       & 60 & 40 \\
TDF-DM        & 50 & 80 \\
\midrule
\textbf{VISTA (\textit{ours})} & \textbf{100} & \textbf{0} \\ 
\bottomrule
\end{tabular}
}
\end{table}%

\subsubsection{Multi-Object Training}

As shown in Table~\ref{table:multi_bottle_capping}, the performance of the DP3-Wrist does not improve despite being exposed to all cap sizes during training; issues such as over-gripping and bottle dislocation due to asymmetric contact persist. Furthermore, for medium-sized caps (4\,cm), a primary failure factor is severe axis misalignment during the secondary repositioning process after an initial unsuccessful attempt, rendering subsequent twisting actions completely ineffective. In comparison, VISTA-Policy maintains smoother overall motions and successfully performs secondary attempts to open the caps. This multi-stage execution capacity further supports \textbf{Q1} and \textbf{Q2}, proving that the distinct and robust contact signals provided by the VDF guide the model to overcome minor noise disturbances and execute intended actions.

\subsection{Calligraphy Writing}
\subsubsection{Main Results and Stroke Quality Evaluation}
This task demands stringent $z$-axis compliance and real-time contact perception. Quantitative results are reported in Table~\ref{table:calligraphy_results}. While the DP3-Wrist achieves a competitive HSR, it struggles to discern the transition from the descending phase to the writing phase. During deployment, the brush frequently oscillates vertically before contacting the paper or descends abruptly, causing excessive mechanical pressing. We attribute this failure to the absence of the VDF---an explicit physical indicator of the writing phase---as sparse and cluttered PC modalities cannot effectively guide fine-grained manipulation. Furthermore, the standard DP3 frequently skips the tool-retrieval phase (causing immediate task failure) and exhibits unstable gripping, although its paper-contact rate marginally exceeds that of DP3-Wrist. We infer that the wrist-view PC helps verify grasping during early trajectories, whereas the global-only perspective of standard DP3 fails to reliably guide the complete motion sequence. In contrast, VISTA-Policy demonstrates robust adaptability across various brush types and surface heights. Once the VDF registers sufficient $z$-axis deformation, the policy smoothly transitions into the horizontal writing phase. This precise phase transition, contrasted with baseline failure modes, further validates \textbf{Q1} and \textbf{Q2}. Additionally, the DP3-Wrist-40 still lags behind VISTA-Policy trained on 20 demonstrations (Fig.~\ref{fig:calligraphy_metrics}), indicating that simply expanding the sample size cannot resolve underlying perceptual bottlenecks.

\begin{figure*}[t] 
  \centering
  \includegraphics[width=0.98\textwidth]{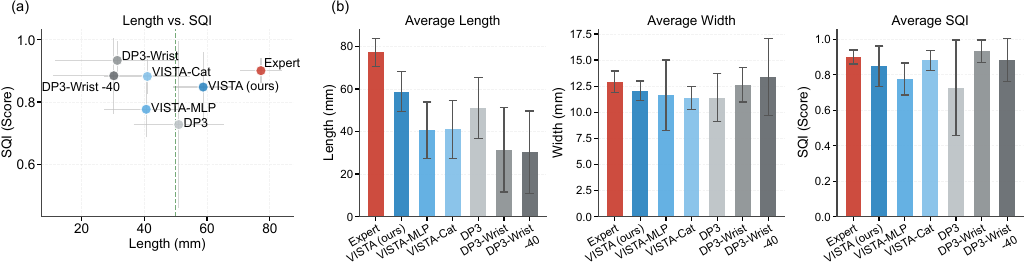} 
  
  \caption{Evaluation of calligraphy stroke quality and metrics: (a) Scatter plot of Stroke Quality Index (SQI) versus stroke length across different methods to reveal degenerate cases with high SQI but insufficient stroke length; (b) Comparison of average calligraphy stroke metrics (length, width, and SQI) with error bars across VISTA-Policy and the baselines.}
  \label{fig:calligraphy_stroke} 
\end{figure*}

\begin{table}[htbp] 
\centering
\caption{Quantitative Results of the Calligraphy Writing Experiment}
\label{table:calligraphy_results}
\renewcommand{\arraystretch}{1.1}
\resizebox{\columnwidth}{!}{ 
\setlength{\tabcolsep}{18pt} 
\begin{tabular}{lcc}
\toprule
\textbf{Method} & \textbf{HSR (\%)} $\uparrow$ & \textbf{WSR (\%)} $\uparrow$ \\
\midrule
DP3             & 50  & 40 \\
DP3-Wrist  & 90  & 40 \\
VISTA-Cat       & 70  & 30 \\
VISTA-MLP       & 100 & 60 \\
TDF-DM        & 80  & 20 \\
\midrule
\textbf{VISTA (\textit{ours})} & \textbf{100} & \textbf{90} \\
\bottomrule
\end{tabular}
}
\end{table}%

We further evaluate calligraphy stroke quality. First, the fidelity distribution chart (Fig.~\ref{fig:calligraphy_stroke}(a)) maps the correlation between $SQI$ (performance) and stroke length (completion), with cross-error bars indicating variance bounds. As the preceding analysis shows, the DP3-Wrist and DP3-Wrist-40 frequently exhibit stationary vertical oscillations, typically leaving only ink dots or extremely short segments. These dot-like strokes should not be regarded as high-quality writing, yet their solid appearance can artificially inflate the computed stroke fullness metric. Consequently, these groups cluster in the upper-left quadrant---featuring high $SQI$ but extremely low lengths---with extensive variance bounds. While standard DP3 performs adequately in stroke length, its $SQI$ is low with extensive variance, indicating highly unstable performance. Conversely, the VISTA-Policy distribution (blue markers) shifts distinctly toward the upper-right quadrant, closely approaching the expert demonstrations (red markers), strongly corroborating its superior writing fidelity. Finally, the performance metric bar charts (Fig.~\ref{fig:calligraphy_stroke}(b)) demonstrate that VISTA-Policy closely tracks the training set averages with minimal variance, exhibiting \textbf{superior execution quality and stability} compared to all baselines and ablation groups.

\begin{figure}[t] 

    \centering

    \includegraphics[width=\linewidth]{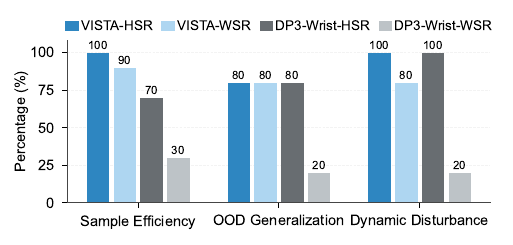} 

    \caption{Quantitative results of sample efficiency and robustness evaluation in \textit{Calligraphy Writing}.} 

    \label{fig:calligraphy_metrics} 

\end{figure}

\subsubsection{OOD Generalization and Dynamic Disturbance Robustness}
In OOD height evaluations, the DP3-Wrist exerts excessive pressing behavior when deployed beyond the trained height range, leading to writing failures; in lower-height configurations, it fails to command sufficient downward extension, causing the brush to merely sweep above the paper surface without contact. Conversely, VISTA-Policy substantially outperforms the baseline, enabling reliable paper contact and short-stroke writing even at an extreme lower bound of 15\,cm. This superior OOD generalization provides an affirmative answer to \textbf{Q2}. Furthermore, during dynamic-disturbance tests where the surface height changes online during execution, VISTA-Policy successfully modulates its vertical action adaptively to complete the effective stroke. In contrast, the control stability of the DP3-Wrist collapses immediately upon environmental height transitions. Quantitative results are illustrated in Fig.~\ref{fig:calligraphy_metrics}, highlighting the robustness of VISTA-Policy.

\subsection{Ablation Studies} 

Across all tasks, the ablation variants (VISTA-Cat and VISTA-MLP) exhibit performance degradation compared to the full VISTA-Policy framework. In \textit{Cross-Scale Object Grasping}, VISTA-MLP prematurely triggers the lifting phase upon minor deformation during initial contact, causing frequent drops during transport. Meanwhile, VISTA-Cat exhibits widespread instability, including descent misalignment and gripper closure stuttering. In \textit{Cap Unscrewing}, VISTA-Cat executes sluggishly and fails to perform re-twisting maneuvers. In the \textit{Calligraphy Writing} task, VISTA-Cat is highly susceptible to pre-contact interference, exhibiting repetitive vertical oscillations and hesitation before tool retrieval. While VISTA-MLP holds the tool marginally better, it suffers from severe vertical bouncing during writing, degrading stroke quality.

In summary, VISTA-Cat yields heavily stuttered motions and prolonged execution times. Mechanistically, the direct concatenation injects background and visual tracking noise into the policy stream, disrupting action causality and often causing total task failure. For VISTA-MLP, the simple MLP encoder cannot accurately capture the spatio-temporal characteristics of the VDF, fundamentally limiting its ability to augment manipulation performance.

\subsection{Interpretability Analysis on Feature Effectiveness}
\begin{figure}[t] 

    \centering

    \includegraphics[width=\linewidth]{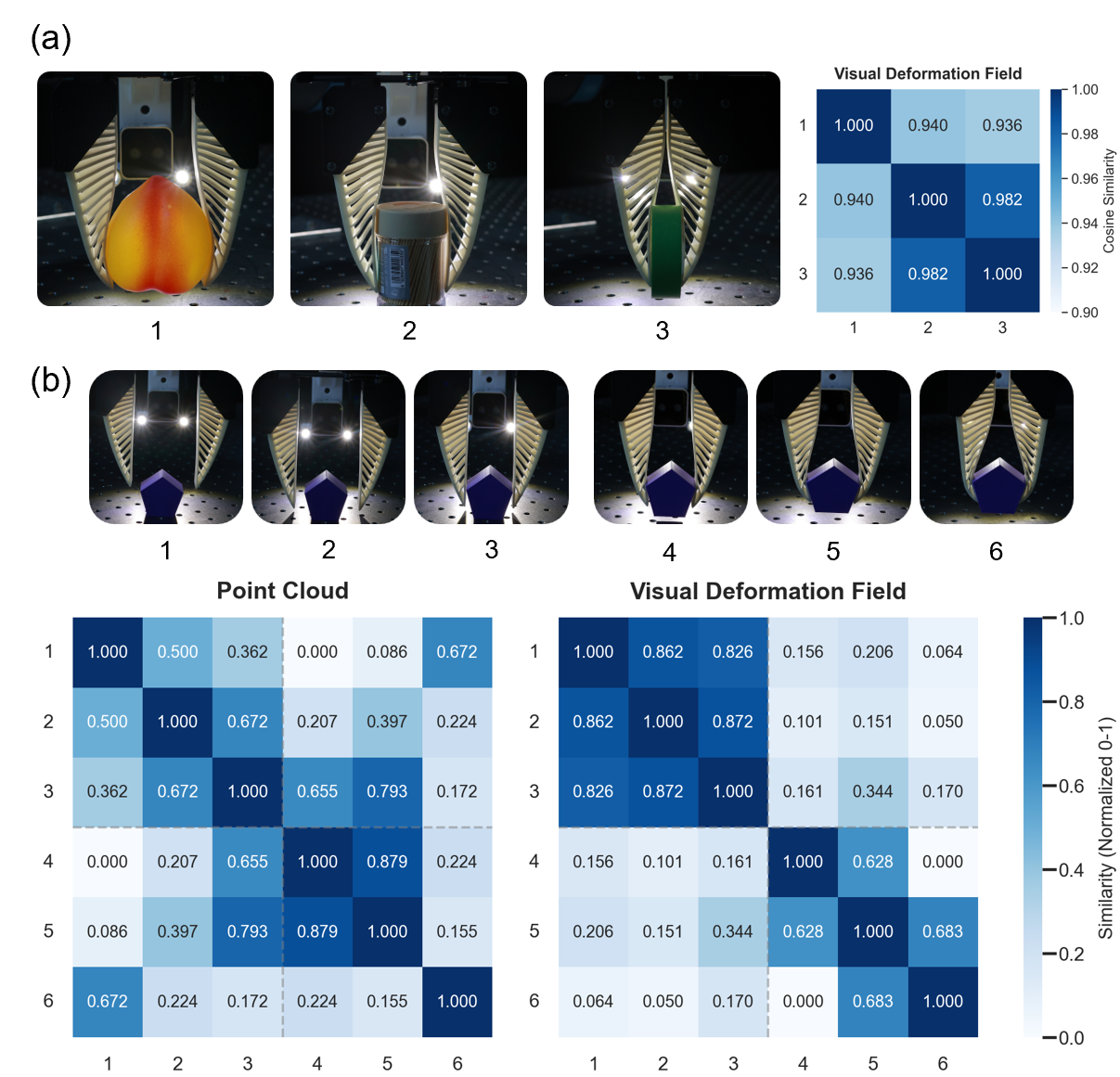} 

\caption{Feature similarity analysis for validating contact representations. 
(a) Feature consistency across objects with different scales and geometries under similar local contact states. 
(b) Comparison of contact-phase separation between the point-cloud features and VDF features across sequential non-contact frames (1--3) and contact frames (4--6).}

    \label{fig:fig4} 

\end{figure}

To evaluate the efficacy of the VDF and its dedicated encoder, we design a feature similarity analysis experiment. As illustrated in Fig.~\ref{fig:fig4}(a), although the three objects exhibit distinct global geometries and scales, their localized contact states---reflected by the degree of gripper depression---remain largely consistent. The corresponding cosine similarity matrix reveals consistently high pairwise similarity among the extracted feature vectors, confirming that our paradigm reliably maps diverse object instances into a \textbf{unified, physically invariant contact representation} (\textbf{Q1}).

Furthermore, Fig.~\ref{fig:fig4}(b) validates the distinct interaction phases during a grasping sequence by tracking six sequential timesteps: the first three correspond to non-contact states, and the latter three capture progressive tightening under contact states. The cosine similarity matrices demonstrate that features from the baseline DP3 encoder produce cluttered correlations that fail to capture these state transitions. Conversely, our VDF similarity matrix exhibits a distinct block-diagonal topology: Non-contact features remain highly similar, while contact-state features evolve progressively as deformation increases. This clear phase separation indicates that the VDF encoder effectively reduces \textbf{contact-state ambiguity} and delivers clean, meaningful contact representations for downstream policies, fully validating \textbf{Q1}.

\subsection{Discussion and Practical Implications}

\textbf{Mechanistic Insights into Pure-Vision Failures:} Pure-vision baselines frequently exhibit failure modes such as premature lifting and severe over-gripping, as geometric observations alone provide limited physical closed-loop feedback and may entangle subtle contact cues with object geometry. By explicitly extracting the VDF feedback, VISTA-Policy provides a structured physical prior that enforces genuine interaction causality, driving its superior data efficiency and generalization (answering \textbf{Q1, Q2}). Detailed mechanistic analyses are provided in the Supplementary Material.

\textbf{Limitations of Tactile Hardware Baselines:} The TDF-DM frequently suffers from persistent stuttering and scraping and triggers protective stops across tasks, leading to suboptimal performance: a high 80\% DR in \textit{Cap Unscrewing} and a mere 20\% SR in \textit{Calligraphy Writing}. Conversely, VISTA-Policy comprehensively outperforms TDF-DM across all metrics, including safety (answering \textbf{Q2, Q3}). Detailed mechanistic analysis is provided in the Supplementary Material.

\textbf{Deployment Feasibility and Cost-Benefit Analysis:} Beyond superior control and generalization, VISTA-Policy demonstrates substantial cost-effectiveness, reliability, and operational lifespan. The compliant gripper costs merely \$77 and successfully withstands over 1,000 high-intensity contact-rich trials without structural degradation (details in the Supplementary Material).

\section{Conclusion and Limitations}

This paper proposes VISTA-Policy, a novel IL paradigm that extracts the observable deformation of a passive compliant gripper as visuo-physical feedback. By integrating real-time VDF extraction, spatial energy denoising, and a deformation-augmented policy network with relative incremental gripper actions, VISTA-Policy converts explicit physical interactions into structured contact signals. Extensive evaluations on \textit{Cross-Scale Object Grasping}, \textit{Cap Unscrewing}, and \textit{Calligraphy Writing} demonstrate that VISTA-Policy substantially outperforms the strong pure-vision baseline and the tactile baseline, with strong OOD generalization and robustness. Experimental results further show that simply increasing demonstration data is insufficient for policies to capture the essence of contact interactions. In contrast, embedding VDF as a low-cost physical prior into policy learning provides an effective pathway toward general-purpose contact-rich robotic manipulation.

Despite these results, VISTA paradigm still has several limitations for future investigation. First, VISTA relies on observable compliant-gripper deformation; under complete occlusion or extremely low-light conditions, feature extraction may degrade. Incorporating temporal memory or multi-view fusion could further improve robustness in extreme environments. Second, the current deformation representation is mainly used as contact-state feedback. Future work may explore the mapping between deformation patterns, object compliance, and contact forces, extending VISTA from manipulation feedback toward compliance perception and force estimation. Furthermore, integrating this low-cost VDF feedback with VLA may bridge high-level semantic understanding and low-level fine-grained control, enabling broader applications in sorting, delicate packaging, and service robotics.

\bibliographystyle{IEEEtran}
\bibliography{references}

\vfill

\end{document}